\documentclass[runningheads]{llncs}
\usepackage{graphicx}
\usepackage{hyperref}

\usepackage[capitalize]{cleveref}
\usepackage{amssymb}
\usepackage{pifont}
\usepackage{color}
\usepackage{booktabs}

\usepackage{hyperref} 
\hypersetup{
    colorlinks=true,
    citecolor=green
}

\begin{document}
\title{A BERT-Style Self-Supervised Learning CNN for Disease Identification from Retinal Images}
\titlerunning{A BERT-Style Self-Supervised Learning CNN}
%
\author{Xin Li\inst{1}\and
Wenhui Zhu\inst{1} \and
Peijie Qiu\inst{2} \and
Oana M. Dumitrascu\inst{3} \and 
Amal Youssef \inst{3}
Yalin Wang\inst{1} 
}
\institute{School of Computing and Augmented Intelligence, Arizona State University, AZ 85281, USA \and
McKeley School of Engineering, Washington University in St. Louis, St. Louis, MO 63130, USA \and
Department of Neurology, Mayo Clinic, Scottsdale, AZ 85251, USA
}

\maketitle              
\begin{abstract}
In the field of medical imaging, the advent of deep learning, especially the application of convolutional neural networks (CNNs) has revolutionized the analysis and interpretation of medical images. Nevertheless, deep learning methods usually rely on large amounts of labeled data. In medical imaging research, the acquisition of high-quality labels is both expensive and difficult. The introduction of Vision Transformers (ViT) and self-supervised learning provides a pre-training strategy that utilizes abundant unlabeled data, effectively alleviating the label acquisition challenge while broadening the breadth of data utilization. However, ViT's high computational density and substantial demand for computing power, coupled with the lack of localization characteristics of its operations on image patches, limit its efficiency and applicability in many application scenarios. In this study, we employ nn-MobileNet, a lightweight CNN framework, to implement a BERT-style self-supervised learning approach. We pre-train the network on the unlabeled retinal fundus images from the UK Biobank to improve downstream application performance. We validate the results of the pre-trained model on Alzheimer's disease (AD), Parkinson's disease (PD), and various retinal diseases identification. The results show that our approach can significantly improve performance in the downstream tasks. In summary, this study combines the benefits of CNNs with the capabilities of advanced self-supervised learning in handling large-scale unlabeled data, demonstrating the potential of CNNs in the presence of label scarcity.

\end{abstract}

\section{Introduction}
Over the past decades, convolutional neural networks (CNNs), have revolutionized the field of medical imaging by achieving outstanding results in a variety of tasks~\cite{Sarvamangala:CNNSurvey2022,denosing1,denosing2}.
CNNs have achieved excellent results in tasks such as retinal disease detection~\cite{zhang2019canet,han2021rethinking,simonyan2014very}, thanks in large part to their core properties of spatial hierarchy, localization, and translation invariance. These features enable CNNs to capture local visual features such as edges and textures and transform them into higher-level abstract features~\cite{zhang2019canet}. This ability is crucial as it allows for the identification of subtle differences that are critical for disease diagnosis. 

 Deep learning shows great potential in medical image analysis. However, the difficulty of acquiring labels and the cost they demand remains a notorious problem for the field~\cite{zhou2019collaborative}. In recent years, the field of natural language processing (NLP) has made significant progress in self-supervised pre-training models~\cite{qiu2020pre}, such as the self-supervised learning models of bidirectional encoder representations from transformers (BERT)~\cite{devlin2018bert} and generative pre-trained transformers (GPT)~\cite{brown2020language}. Drawing inspiration from the achievements in NLP, the integration of vision transformers (ViT) has catalyzed the development of numerous self-supervised vision models employing masking strategies. This shift marks a significant transition in the vision domain from contrastive learning to generative learning approaches in self-supervised models.
 

Although ViTs bring new approaches to solve problems in the medical imaging field~\cite{yu2021mil,zhou2019collaborative}, the computationally high intensity of their self-attention mechanism and their reliance on large-scale datasets reveal their limitations~\cite{yu2021mil,li2021localvit}. The time and memory complexity of self-attention operations grows quadratically with the input size, making ViTs particularly resource-intensive when processing large amounts of data. Furthermore, compared to traditional CNNs, ViTs exhibit less localization capability in image processing, as their mechanisms primarily target the patch level of images, rather than focusing directly on individual pixels.~\cite{li2021localvit}. The advantages we observed in CNNs, along with the significant benefits of pre-training for self-supervised learning, and the potential shortcomings of ViTs, all stimulate further thought: Is it possible to combine CNNs of good performance with generative self-supervised learning in the BERT style for pre-training, to incorporate the advantages of both?

\begin{figure}[t]
\begin{center}
\includegraphics[width=0.9\textwidth]{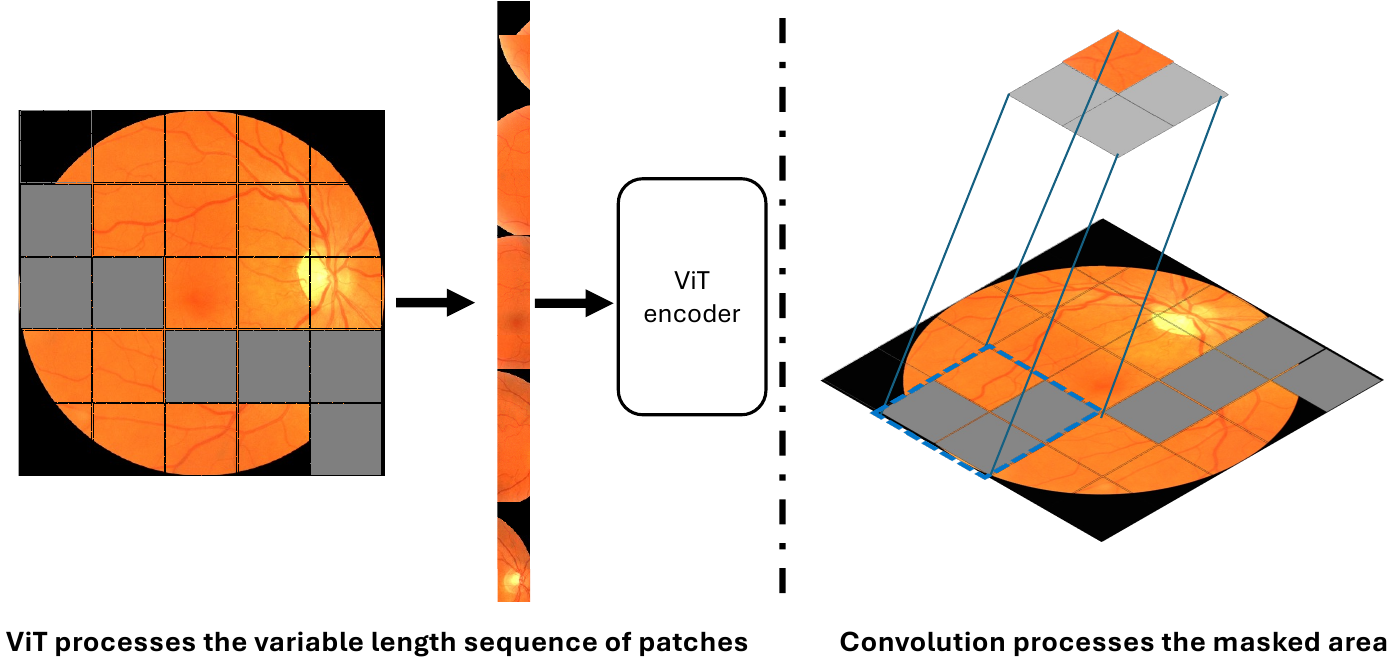}
\label{vit_vs_cnn}
\caption{The ViT on the left panel can process non-masked patches without any changes since it can process variable-length sequences, while the CNN on the right panel cannot skip masks for convolutions. Simply adopting masking in CNN may lead to performance degradation. This work adopts a novel solution to address this problem.}
\end{center}
\end{figure}

Our study reveals that initial endeavors~\cite{pathak2016context,zhang2017split} to train CNNs using a masking strategy for self-supervised learning did not achieve competitive results.
The inherent architectural distinctions between CNNs and ViTs hinder the direct substitution of CNNs for ViTs in image processing tasks (Fig.~\ref{vit_vs_cnn}). Unlike ViTs, which adeptly handle variable-length sequences of patches without the need for adjusting masking regions, CNNs cannot bypass masked areas during convolution operations. This limitation results in inconsistent outcomes and diminished performance in CNNs compared to ViTs.
Recently, some researchers have addressed this masking problem by introducing sparse convolution~\cite{liu2015sparse}, such as the proposal of SparK~\cite{tian2023designing} and ConvNextV2~\cite{woo2023convnext}. With the sparse convolution, this study introduces a BERT-style self-supervised learning framework for CNNs, utilizing nn-MobileNet~\cite{zhunnmobilenet} as the CNN backbone, a lightweight architecture derived from MobileNetV2~\cite{sandler2018mobilenetv2}. Leveraging  178,803 unlabeled retinal images from the UK Biobank~\cite{UKBiobank-plos2015}, we conduct self-supervised pre-training. This pre-trained model is then evaluated on downstream tasks to assess its performance. Our findings highlight the significant potential of CNNs, particularly in scenarios characterized by a lack of labeled data.


The main contributions of this paper include: \textbf{1).} By adopting the sparse convolution, we propose a novel BERT-style self-supervised learning CNN to enrich self-supervised approaches. It is universally applicable across a broad spectrum of medical imaging research. \textbf{2).} 
Our method harmoniously integrates with a CNN model~\cite{zhunnmobilenet}, capitalizing on its strengths like precise localization and lower data requirements. \textbf{3).} Our extensive experiments confirm its superiority over several leading supervised~\cite{zhang2019canet,han2021rethinking} and self-supervised~\cite{he2022masked,zhou2023foundation} models across a spectrum of benchmarks, including one~\cite{zhou2023foundation} pre-trained with over 1.6M fundus images.

\section{Method}


\subsection{nn-MobileNet}
\begin{figure}[t]
\centering
\includegraphics[width=0.9\textwidth]{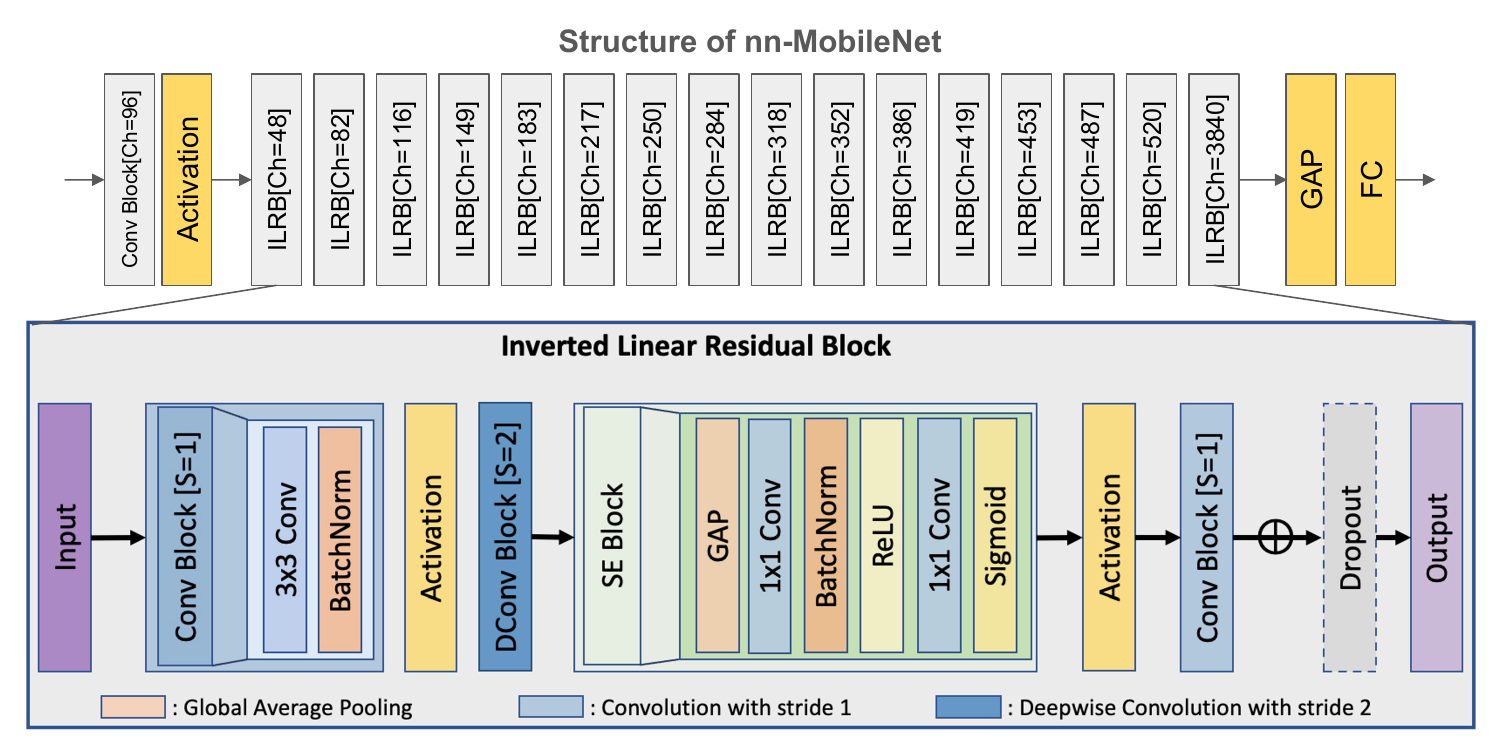}
\caption{The detailed architecture of the nn-MobileNet and its ILRB design. The nn-MobileNet achieved superior results in retinal imaging research~\cite{zhunnmobilenet}. The current self-supervised learning scheme further enhances its performance.}
\label{nnmobileNet}
\end{figure}

Our backbone is a recently developed CNN model nn-MobileNet~\cite{zhunnmobilenet} (Fig.~\ref{nnmobileNet}). Based on the MobileNetV2~\cite{sandler2018mobilenetv2}, it makes the following architectural innovations.  

\textbf{Channel Configuration.}
The nn-MobileNet modifies the order of channel configurations for the inverted linear residual bottleneck (ILRB) in the network. This strategy aims to improve network performance by leveraging the significant impact of channel configurations~\cite{liu2022convnet}.

\textbf{Data Augmentation.}
Compared with conventional retinal imaging methods~\cite{zhang2019canet}, the nn-MobileNet employs a heavy data augmentation strategy, including image cropping, flipping, contrast adjustment, brightness adjustment, and sharpening. Experiments have demonstrated that this heavy data enhancement has significant benefits for improving system performance~\cite{zhunnmobilenet}.

\textbf{Dropout.}
In order to solve the overfitting problem, the nn-MobileNet attempts to add Dropout modules at various locations within the network to identify the optimal placement for them as shown in Fig. \ref{nnmobileNet}.

\textbf{Activation Functions.}
According to prior studies, smooth variants of the ReLU activation functions can improve the performance. Therefore, through experimental comparisons, ReLU6 is chosen as the activation function for every ILRB in the framework due to its superior performance enhancement.

\textbf{Optimizer.}
By experimenting and comparing current common optimizers (including Adam, AdamW, AdamP, and SGD), AdamP~\cite{heo2020adamp} is empirically chosen.

\subsection{Pre-Training Strategy}
\begin{figure}[t]
\centering
\includegraphics[width=0.9\textwidth]{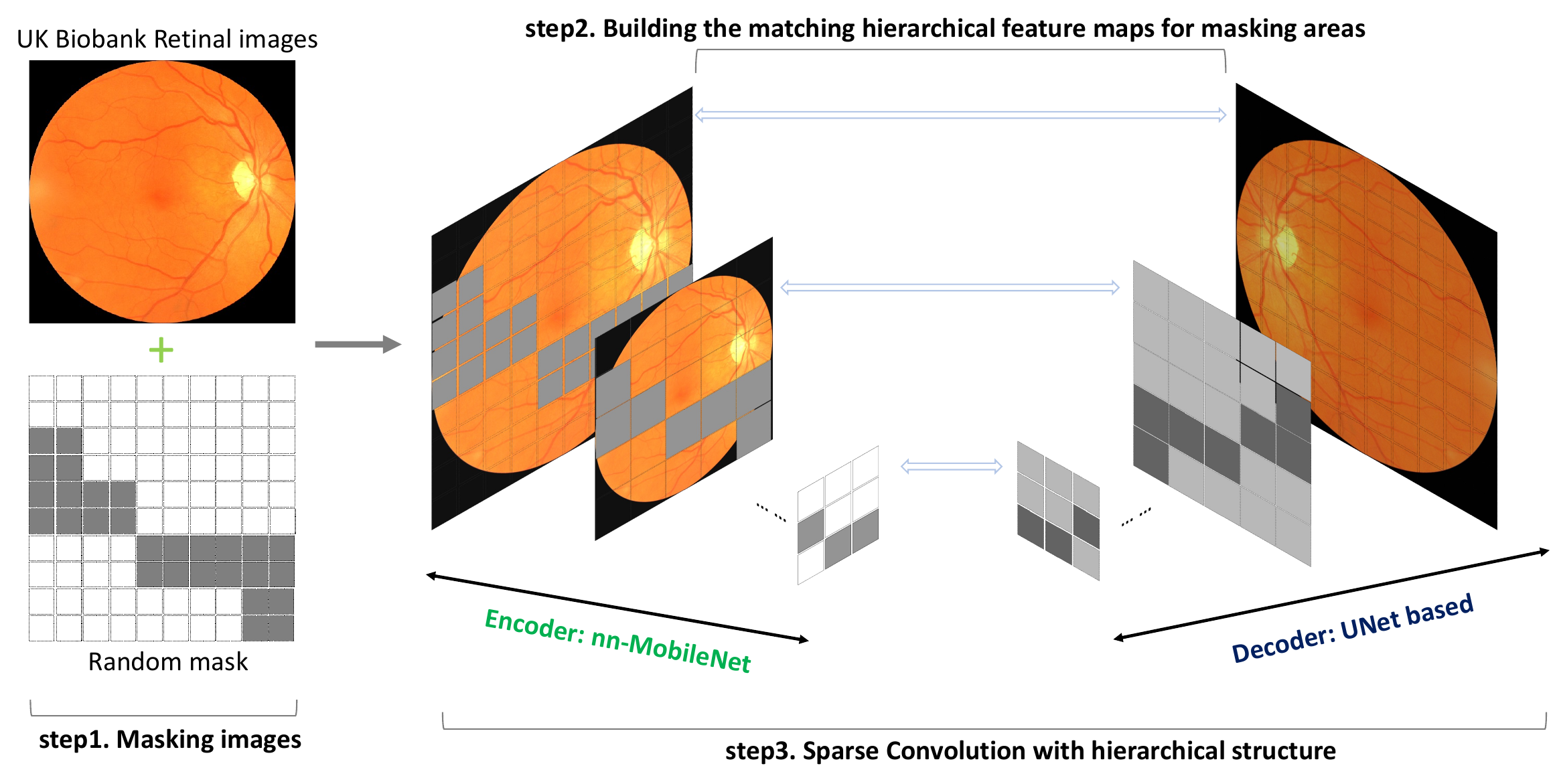}
\caption{Illustration of our pre-training workflow. We start by masking all the images randomly (Step 1). Next, we mask the feature maps adapted to different resolutions for the CNN encoder and decoder (Step 2). Finally, we perform sparse convolution on the masked image and restore the image through the decoder (Step 3).}
\label{pretrain} 
\end{figure}

Fig.~\ref{pretrain} illustrates the workflow of the pre-training strategy, which draws on the basic principles of the BERT architecture. We employ image masking techniques (Step 1) and match hierarchical feature maps for masking areas (Step 2) to construct the loss functions. We use sparse convolution~\cite{liu2015sparse} to integrate the CNN architecture (Step 3). Finally, the learned feature maps are used for downstream applications. Details of this pre-training method are elaborated as follows.

\textbf{BERT Style.}
In the field of computer vision, the masked autoencoder (MAE)~\cite{he2022masked}, similar to the BERT~\cite{devlin2018bert} model from the NLP field, has become one of the most advanced pre-training strategies for self-supervised learning. However, the research based on MAE methods has been mainly based on ViTs. As discussed in the previous section, due to their inherent differences, the encoder-decoder structure of ViTs cannot be directly substituted by CNNs. Until recently, with the successive proposals of SparK~\cite{tian2023designing} and ConvNextV2~\cite{woo2023convnext}, new ideas have emerged for adopting CNNs for the pre-training research with the BERT-style self-supervised learning approach.

\textbf{Hierarchical Structure.}
Hierarchy is widely recognized as the gold standard for vision representation systems. The hierarchical design implemented in various CNNs~\cite{sandler2018mobilenetv2,woo2023convnext} significantly improved their performance. However, this principle is not utilized in ViTs.
In our approach, we adopt SparK~\cite{tian2023designing} to maintain the hierarchy inherent in CNNs and ensure that the neural network can utilize the hierarchical structure for improved representation learning.
  
\textbf{\textit{Encoder:}}
Before performing the convolution operation, we generate feature maps with different resolutions based on the downsampling process of the CNN. For our neural network model nn-MobileNet, a total of five downsamplings are performed on an image with a size of $H*W$, where H,W are the height and width of the input image, respectively, resulting in the generation of the size of feature maps set S with the following:

\[ 
S_i = \left\{ \left( \frac{H}{2^i}, \frac{W}{2^i} \right) \right\},\quad \forall i \in \{1, 2, 3, 4, 5\}. 
\]

\textbf{\textit{Decoder:}}
In the decoding stage, we adopt a lightweight UNet decoder~\cite{ronneberger2015u}, which is characterized by the inclusion of four successive blocks \{$B_1,B_2,B_3,B_4$\} with upsampling layers. The decoder receives the $S'_i$ from the different resolution feature maps $S_i$ and its mask embedding. Furthermore, by applying a projection layer $\phi_i$, the dimensional consistency between the encoder and decoder is ensured. For the smallest feature map, we define: $D_5 = \phi_5(S'_5)$. From this, we derive the rest of the feature maps $D_i$:
\[
D_i = B_i(D_{i+1}) + \phi_i(S'_i), \quad \forall i \in \{4, 3, 2, 1\}.
\]

\textbf{Loss Function:}
Our loss function utilizes a method of comparing the mean squared error (MSE) between the masked portion of the reconstructed images and the corresponding masked region in the original images.

\begin{figure}[t]
\centering
\includegraphics[width=0.9\textwidth]{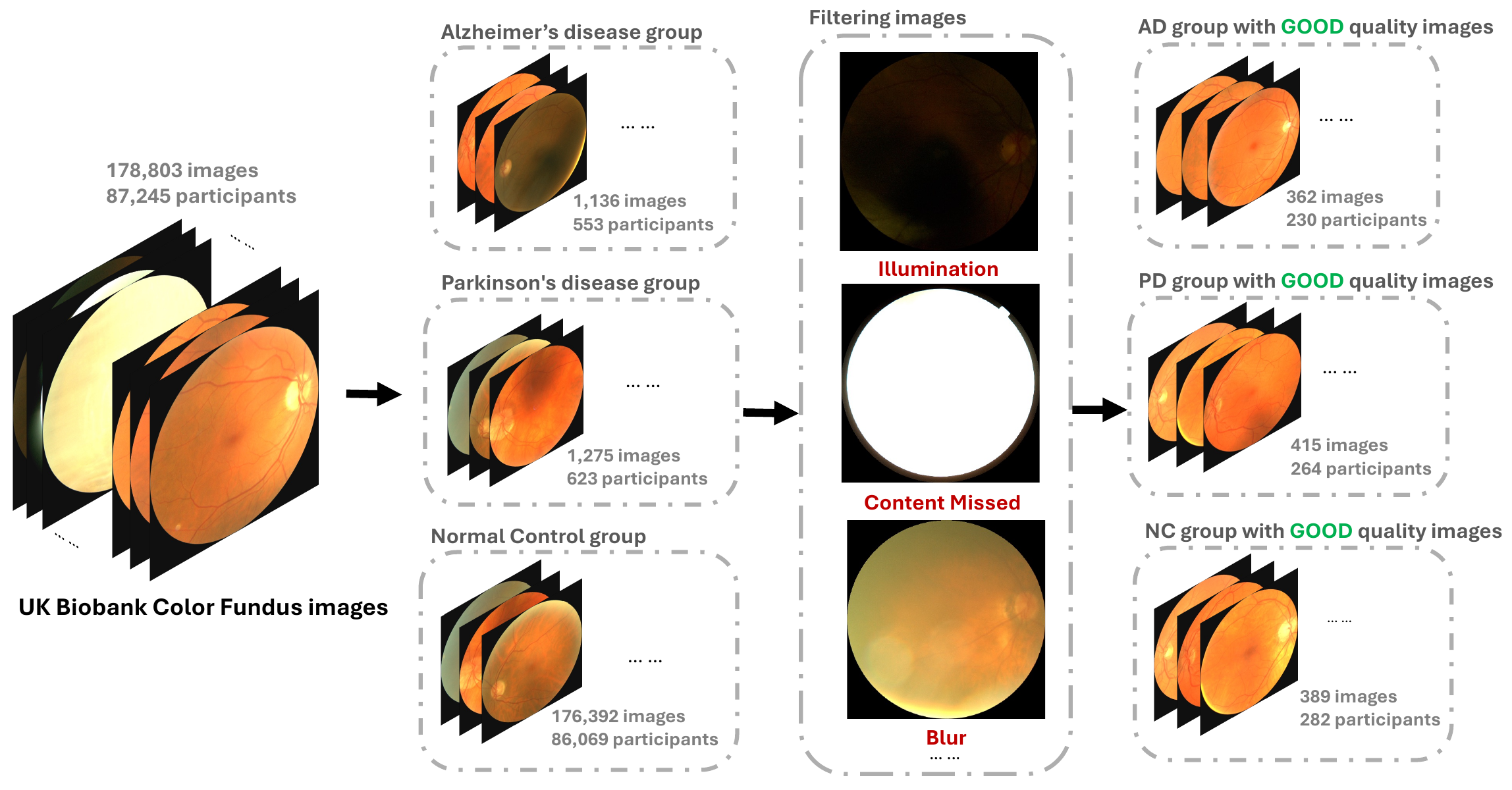}
\caption{Illustration of the quality control pipeline as we filter fundus image data of AD patients, PD patients, and normal control subjects from the UK Biobank dataset.} 
\label{uk_sep}
\end{figure}

\textbf{Sparse Convolution.}
Convolution is widely used in 2D computer vision research, where sliding window operations are usually performed on a grid of pixels. However, it is inappropriate in MAE studies because masked pixels might participate in the convolutions and lead to inconsistent results. Sparse convolution~\cite{liu2015sparse} improves the computation consistency by omitting all the empty voxels and focusing convolution operations on non-zero elements. This work extends the sparse convolution to the self-supervised research by ensuring that convolution operations are conducted exclusively on non-masked pixels. 

\section{Experimental Results}

\textbf{Pre-training on the UK Biobank dataset.}
Our study uses 178,803 unlabeled color retinal fundus images from the UK Biobank~\cite{UKBiobank-plos2015}, involving 87,245 participants, for the self-supervised pre-training. During the pre-training, we first randomly mask $60\%$ of the input image, where the size of each mask is equal to the: ($\frac{H}{D}$, $\frac{W}{D}$), with D being the downsampling ratio of the network. Then, we process only the unmasked visible regions of the input image. In our network, the size of the output image is $(224,224)$ and the downsampling ratio is $32$, making our mask size to be $(7,7)$. We further use the feature maps learned from pre-training module for classification studies. 

\textbf{Benchmarks.} For experiments, we randomly divided the data into training and validation sets in the ratio of 8:2. We employed 5-fold stratified cross-validation on the dataset. The evaluation metrics used to assess the improvement in network performance after pre-training include Quadratic-weighted Kappa (Kappa), AUC (Area Under the Receiver Operating Characteristic), F1 score, and accuracy. Our comparison methods encompass CANet~\cite{zhang2019canet}, VGG11~\cite{simonyan2014very}, Rexnet~\cite{han2021rethinking} and ViT-based approaches, including MAE~\cite{he2022masked}, RETFound~\cite{zhou2023foundation}, and a pre-trained MAE model based on the UK Biobank.

\textbf{Alzheimer's Disease (AD) and Parkinson's Disease (PD) Identification.}
Our quality control selects 362 fundus images of AD patients from the UK Biobank, which comprises 169 images from the left eye and 193 images from the right eye among 230 patients. Additionally, we match 389 images from participants without AD, comprising 170 images from the left eye and 219 images from the right eye among 282 participants, as a reference group. Similarly, we select 415 fundus images of PD patients from the UK Biobank, comprising 197 images from the left eye and 218 images from the right eye among 264 patients. Fig.~\ref{uk_sep} illustrates our quality control procedure. We filter the images exhibiting any of the conditions: blur, low contrast, poor illumination, and artifacts. Although we use the fundus images derived from patients with a diagnosis of AD and PD from the UK Biobank, we argue that our pre-training \textbf{1)} does not use the image labels, and \textbf{2)} primarily focuses on learning mask regions to build the feature maps. Our practice is consistent with common self-supervised approaches~\cite{zhou2023foundation}.

\textbf{MICCAI Myopic Maculopathy Analysis Challenge (MMAC) 2023 - Classification of Myopic Maculopathy~\cite{challenge}.}
The challenge contains 1143 fundus images categorized into four myopic maculopathy grades. There are 404 images for grade 0, 412 images for grade 1, 224 images for grade 2, 60 images for grade 3, and 43 images for grade 4. The nn-MobileNet achieved the third rank in the MICCAI MMAC~\cite{challenge}, which we utilize to demonstrate the advantage of our pre-training strategy.

\begin{table}[t]
\centering
\small 
\caption{The experimental result comparison of the proposed pre-trained model with other supervised and self-supervised networks on the AD/PD identification studies. Our results are either in par with or outperform these state-of-the-art methods.}

\label{tab:baseline}
\resizebox{0.7\textwidth}{!}{

       \begin{tabular}{lcccccc}
         \toprule
                 \multicolumn{1}{c}{}  & \multicolumn{3}{c}{AD} & \multicolumn{3}{c}{PD}\\

              \cmidrule(lr){2-4}  \cmidrule(lr){5-7}   
          Method  & ACC& AUC& Kappa& ACC& AUC& Kappa \\
         \midrule
         CANet~\cite{zhang2019canet} & 0.7 & 0.7391 & 0.3981 & 0.6273 & 0.6758 & 0.2557       \\
         VGG11~\cite{simonyan2014very} & 0.7467 & 0.8114 & 0.4898      & 0.6211   & 0.589 & 0.2447       \\
         Rexnet~\cite{han2021rethinking} & 0.9933 & 1.0 & 0.9866 & 0.7640 & 0.8974 & 0.534       \\
         MAE~\cite{he2022masked} & 0.9933 & 1.0 & 0.9866 & 0.8385 & 0.9132 & 0.677       \\
         MAE+UK Biobank\textsuperscript{*} & 0.9933 & 1.0 & 0.9866 & 0.8696 & 0.9411 & 0.7392       \\
         RETFound~\cite{zhou2023foundation} & 0.9933 & 1.0 & 0.9866 & 0.9193 & 0.9768 & 0.8387       \\
         nn-MobileNet~\cite{zhunnmobilenet} & 0.9933 & 1.0 & 0.9866 & 0.9876 & 0.9993 & 0.9752       \\

         \cline{1-7}
         Ours & \textbf{0.9933} & \textbf{1.0} & \textbf{0.9866} & \textbf{0.9938} & \textbf{0.9993} & \textbf{0.9876}      \\        

         \bottomrule

     \end{tabular}
}

\textsuperscript{*}We continued pre-training with the UK Biobank based on MAE results.~\cite{he2022masked}.


\end{table}

\textbf{Results:} 
Our research method exhibited strong performance on the AD and PD tasks (Table~\ref{tab:baseline}). On the AD dataset, we achieved good results with 99.33\% correctness, a kappa score of 0.9866, and an AUC of 0.9997. On the PD dataset, our method also achieved excellent results with 99.38\% correctness, a kappa score of 0.9876, and an AUC of 0.9993. Our results were either on par with or outperform other state-of-the-art methods~\cite{zhou2023foundation,he2022masked}. It is worth noting the RETFound model~\cite{zhou2023foundation} was pre-trained with over 1.6M retinal images, while ours was trained with 176K fundus images. It is unsurprising since, for the same accuracy level, the CNN usually requires far fewer training data than the ViT~\cite{dosovitskiy2020image}. 

\begin{figure}[t]
\centering
\includegraphics[width=0.8\textwidth]{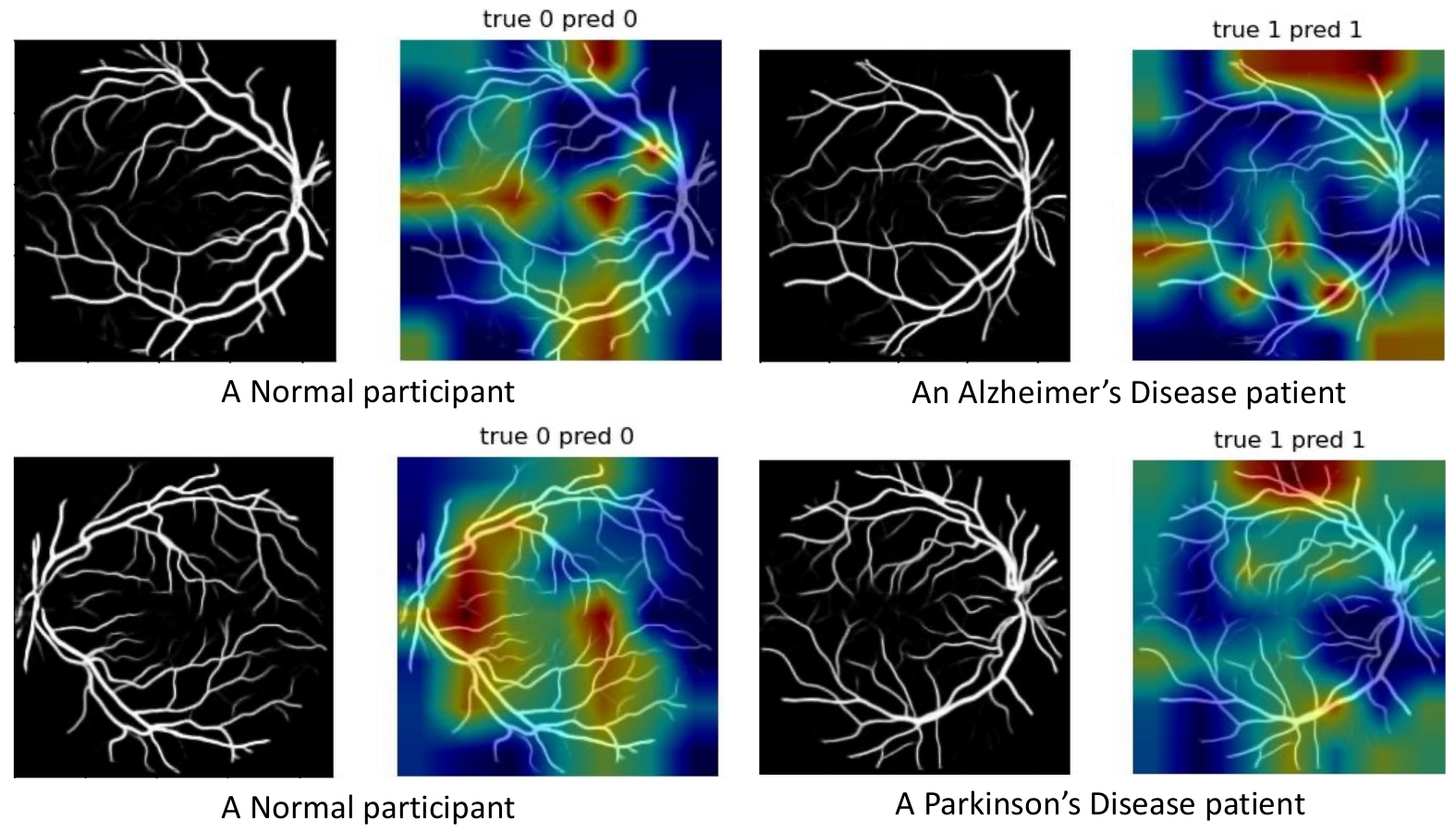}
\caption{Heat maps for AD (the first row) and PD (the second row) demonstrate our work achieves valuable biomarkers consistent with prior research~\cite{Dumitrascu:Cells2021}. } 
\label{heatmap}
\end{figure}

In order to analyze the potential biomarkers, we use the UNet for vessel segmentation to better visualize vessel features~\cite{ronneberger2015u,zhuang2018laddernet}. With them, we generate attention heatmaps with the Grad-CAM~\cite{selvaraju2017grad,jacobgilpytorchcam} (Fig.~\ref{heatmap}). Our results are consistent with the previous AD research~\cite{Dumitrascu:Cells2021}, indicating retinal blood vessel branches with tortuosity change as potential identifiers of AD.

Through experiments on the MMAC dataset (Table.~\ref{tab:rank}), we delved into the impact of pre-training on the performance of the nn-MobileNet. The experimental results revealed that by averaging the results of the 5-fold cross-validation, significant improvements were observed: the Kappa value improved by 0.232, the AUC improved by 0.0039, the Accuracy increased by 0.0238, and the Weighted F1 Score improved by 0.0412. Particularly noteworthy was the substantial enhancement in the Weighted F1 Score, suggesting that pre-training markedly enhances the model's performance in multi-category classification tasks.

\begin{table}[b]
\centering
\caption{Result comparison of the nn-MobileNet~\cite{zhunnmobilenet} and pre-trained nn-MobileNet on the MMAC dataset. The proposed pro-training significantly improves its performance.}
\label{tab:rank}
\resizebox{0.7\textwidth}{!}{

\begin{tabular}{l l l l l l}
\hline
 & epochs &  kappa & AUC & ACC& F1\\
\hline
nn-MobileNet~\cite{zhunnmobilenet} &  300 & 0.8519 & 0.9621 & 0.8105 & 0.7776 \\
pretrained nn-MobileNet &300&  0.8751 & 0.966 & 0.8313 & 0.8188\\
\hline
\bf{Improvement} & & \bf{+0.0232} & \bf{+0.0039} & \bf{+0.0208} & \bf{+0.0412}\\
\hline
\end{tabular}
}
\end{table}



\section{Conclusion and Future Work}
Motivated by advances in NLP, we have successfully combined self-supervised learning with CNNs to break through the computational and data constraints of ViTs through the BERT-style self-supervised training and significantly improve experimental performance. In our future work, we aim to further investigate the advantages of CNN within the realm of self-supervised learning, extending their application to a wider array of medical imaging analyses, including optical coherence tomography (OCT) and magnetic resonance imaging (MRI).

\bibliographystyle{splncs04}
\bibliography{main}  

\end{document}